\documentclass{article}
\usepackage{spconf,amsmath,epsfig}
\usepackage{cite}
\usepackage{graphicx}
\usepackage{amsmath}
\usepackage{multirow}
\usepackage{hyperref}
\usepackage{gensymb}
\usepackage{textcomp}
\usepackage[absolute,showboxes]{textpos}

\TPMargin{5pt}
\title{Amplitude Scintillation Forecasting Using Bagged Trees}
\name{Abdollah~Masoud~Darya, Aisha~Abdulla~Al-Owais, Muhammad~Mubasshir~Shaikh, and~Ilias~Fernini}
\address{Sharjah Academy for Astronomy, Space Sciences, and Technology\\
University of Sharjah, Sharjah, UAE}
\newcommand{\copyrightstatement}{
    \begin{textblock}{0.84}(0.08,0.93)    % tweak here: {box width}(leftposition, rightposition)
         \noindent
         \footnotesize
         \copyright  2022 IEEE.  Personal use of this material is permitted.  Permission from IEEE must be obtained for all other uses, in any current or future media, including reprinting/republishing this material for advertising or promotional purposes, creating new collective works, for resale or redistribution to servers or lists, or reuse of any copyrighted component of this work in other works. This paper was presented at IGARSS 2022. doi: 10.1109/IGARSS46834.2022.9883380
    \end{textblock}
}

\begin{document}
\copyrightstatement
%\ninept
%
\maketitle
\begin{abstract}
Electron density irregularities present within the ionosphere induce significant fluctuations in global navigation satellite system (GNSS) signals. Fluctuations in signal power are referred to as amplitude scintillation and can be monitored through the S4 index. Forecasting the severity of amplitude scintillation based on historical S4 index data is beneficial when real-time data is unavailable. In this work, we study the possibility of using historical data from a single GPS scintillation monitoring receiver to train a machine learning (ML) model to forecast the severity of amplitude scintillation, either weak, moderate, or severe, with respect to temporal and spatial parameters. Six different ML models were evaluated and the bagged trees model was the most accurate among them, achieving a forecasting accuracy of $81\%$ using a balanced dataset, and $97\%$ using an imbalanced dataset.
\end{abstract}
\section{Introduction}
\label{sec:intro}

Scintillation of global navigation satellite system (GNSS) signals is caused by electron density irregularities that are present within the ionosphere \cite{crane1977ionospheric}. Ionospheric scintillations induce significant fluctuations in global navigation satellite systems (GNSS) signals. These fluctuations include variations in the signal-to-noise ratio (SNR), a phenomenon known as amplitude scintillation. Amplitude scintillation may lead to degradation of the quality of service provided by GNSS satellites to users. A typical measure of amplitude scintillation is the S4 index, which is defined as the ratio of the standard deviation of the signal power to the average signal power \cite{yeh2019superposition}. Scintillation monitoring receivers typically provide the S4 index as they continuously monitor the SNR of GNSS signals.\par
As ionospheric scintillations have varying effects based on the geographical location of the user, knowledge of the spatial distribution of scintillation-causing ionospheric irregularities would be very useful. Furthermore, the ability to forecast the severity of these scintillations based on historical data would be advantageous when real-time data is unavailable. Motivated by the points mentioned earlier, in this work, we study the possibility of using historical data from a single GPS scintillation receiver to train a machine learning (ML) model to forecast the severity of amplitude scintillation with respect to temporal and spatial parameters.\par

\subsection{Related Works and Contribution}

Several works have utilized ML to predict or forecast amplitude scintillation, such as \cite{rezende2010survey,linty2018detection,savas2019impact,dey2021automatic,atabati2021ionospheric}. Out of these, the most recent and related to our work are \cite{dey2021automatic,atabati2021ionospheric}. In \cite{dey2021automatic}, the issue of classifying ionospheric amplitude scintillation was addressed by attempting to solve a binary classification problem by considering two levels of scintillation (scintillation or no scintillation). Their best performing classifier was the XGBoost method, which achieved an accuracy of $99.76\%$ for testing. However, they did not consider the spatial allocation of amplitude scintillation even though the variations of amplitude scintillation depend heavily upon spatial parameters. Thus, by not mapping the scintillation values, it would be difficult to localize the predicted scintillation. Furthermore, different scintillation intensities have varying effects on ground users, and thus, unlike \cite{dey2021automatic}, we consider the problem as a 3-class classification problem, by considering weak, moderate, and severe levels of scintillation.\par
In \cite{atabati2021ionospheric}, the use of neural networks for the prediction of S4 values was proposed. The problem was addressed as a regression analysis problem. Furthermore, they included different parameters related to space weather such as F10.7, sunspot number, and peak electron density height in the prediction process. Their designed model was able to predict daily ionospheric scintillations with $81\%$ accuracy. However, like \cite{dey2021automatic} they did not consider the effect of spatial parameters on the predicted scintillation values. Furthermore, their utilization of F2 peak electron density height and F2 vertical drift velocity necessitates the use of ionosonde data, which is less common compared to scintillation monitoring receivers.\par
This work attempts to solve the problem of forecasting the severity of spatially and temporally allocated amplitude scintillation values using bagged trees. This is done by solving a 3-class classification problem using ML methods, to predict weak, moderate, and severe scintillation. The S4 data used in this work was retrieved from a scintillation monitoring receiver at the McMurdo Station in Antarctica, corresponding to a period of four years.\par

\section{Data and Methodology}

% \begin{figure}[htb]
% \center
% \includegraphics[width=\columnwidth]{IPPTracks.jpg} 
% \caption{The red star represents the location of the scintillation receiver location at McMurdo station. The black dots represent the location of the IPP of all observations from the imbalanced dataset.}
% \label{fig1}
% \end{figure}

In this work, we use 1-minute observations of S4 from a GPS scintillation monitoring receiver at McMurdo station ($-77.83\degree$, $166.66\degree$) in Antarctica, from 18-Jan-2011 to 14-Nov-2014. These observations are freely available at \url{http://cedar.openmadrigal.org/}. To map the S4 values to a geographical location, the thin-shell ionospheric model was assumed at an altitude of 350 km. The point at which the GNSS signal intersects with the thin shell is known as the ionospheric pierce point (IPP). The IPP is mapped into latitude and longitude coordinates. The observation data was provided as Ionospheric Scintillation Monitoring Records (ISMR) files. We extracted the time, satellite elevation, latitude and longitude coordinates of the IPP, and S4 values from these files.\par 
One of the most important steps in ML is the preparation and preprocessing of data to improve the learning ability of ML models. For preprocessing, several steps were followed, as shown below:
\begin{enumerate}
  \item Set an elevation cutoff of 20 degrees, to eliminate the effects of multi-path, and removed any $S4<0$ values.
  \item Removed discontinuity in longitude coordinates by adding $360\degree$ to negative longitude values.
  \item At this point the total number of observations was $12\,377\,607$, $67.91\%$ of which were below $0.05$. We assumed S4 values below $0.05$ to represent no scintillation and they were therefore discarded. This reduced the total number of observations to $3\,972\,227$. This served two main purposes. First, removing this large number of S4 observations under $0.05$ would prevent the models from being biased toward very low S4 values. Second, the number of observations would be more manageable, and take considerably less time to train and evaluate.
  \item Added three commonly used solar indices to represent solar activity. These are the daily averaged Planetary Kp-index (KP), Sunspot Number (SSN), and F10.7 index. These indices were obtained from \url{https://omniweb.gsfc.nasa.gov/form/dx1.html}. Next, observations with missing indices, i.e., F10.7 $=999$, were discarded. Furthermore, we did not include parameters such as ionospheric vertical drift velocity and peak electron density height \cite{rezende2010survey,atabati2021ionospheric} to eliminate dependence on ionosonde data.
  \item The data was then divided into three classes. Class 1 represents $S4<0.2$, class 2 represents $0.2\leq S4<0.3$, and class 3 represents $S4\geq 0.3$. These classes correspond to weak, moderate, and severe amplitude scintillation respectively.
  \item For classes 1, 2, and 3, the number of observations were $3\,788\,881$, $157\,163$, and $22\,959$, respectively. Thus, the total imbalanced dataset had a total of $3\,969\,003$ observations. 
  \item We also created a balanced dataset by taking $22\,959$ random samples, corresponding to the smallest class size (class 3), from each class of the imbalanced dataset, adding up to a total of $68\,877$ observations. The balanced dataset should provide another perspective of the performance of ML models as imbalanced data was found to substantially compromise the learning process of ML models \cite{he2009learning}.
\end{enumerate}

After preprocessing, the variables in the dataset consisted of: day of year (DOY), hour of day (HOD), IPP latitude coordinate, IPP longitude coordinate, KP, SSN, F10.7, and S4 class (either 1, 2, or 3). The ML models considered in this work are discussed in the following paragraphs.\par\noindent
\underline{Decision Trees (DT)} consist of nodes, branches, and leaves. It is constructed by splitting data iteratively into branches to form a hierarchical structure. The upper edge of the tree starts from the root node, and the lower edge of the tree consists of leaf nodes. By following a path from the root node to the leaf nodes, the user can obtain a classification from the DT model \cite{sutton2005classification}.\par\noindent
\underline{Naive Bayes (NB)} classifiers assume the attribute values are conditionally independent of each other given the target value or class \cite{chen2020novel}. Thus, in our case, each feature contributes to predicting an S4 value without depending on each other.\par\noindent
\underline{Support Vector Machines (SVM)} create the best decision boundary that can segregate data into their respective classes. This decision boundary is known as the hyper-plane. The data which are closest to the hyper-plane and affect its position, are termed as support vectors, hence the name \cite{steinwart2008support}.\par\noindent
\underline{K-Nearest Neighbors (KNN)} are algorithms that assume that similar data points neighbor each other. Accordingly, it classifies the data to the nearest corresponding class \cite{liao2002use}.\par\noindent
\underline{Boosted Trees} are a type of supervised learning where a model combines the weak learners in an iterative manner and progressively improves the learners. Such method uses a weighted average of the results obtained from applying a prediction model to various samples \cite{sutton2005classification}.\par\noindent
\underline{Bagged Trees} is a method where multiple versions of a classifier, in this case a DT, are generated with each of them generating their own predictions, and the final prediction is the average of all of these individual predictions \cite{sutton2005classification}.

\begin{table*}
\centering
\caption{Results of the unoptimized ML models using the balanced and imbalanced datasets.}
\label{tab1}
\begin{tabular}{|l|l|c|c|}
\hline
Model & Parameters & \multicolumn{1}{l|}{Balanced Dataset Accuracy} & \multicolumn{1}{l|}{Imbalanced Dataset Accuracy} \\ \hline
Decision Tree & \begin{tabular}[c]{@{}l@{}}Maximum number of splits = 100\\ Split criterion = Gini's Diversity Index\end{tabular} & $67.8\%$ & $95.5\%$ \\ \hline
Naive Bayes & Kernel type = Gaussian & $56.1\%$ & $95.5\%$ \\ \hline
SVM & \begin{tabular}[c]{@{}l@{}}Kernel function = Gaussian\\ Kernel scale = 0.66\end{tabular} & $64.1\%$ & $95.5\%$ \\ \hline
KNN & \begin{tabular}[c]{@{}l@{}}Preset = Weighted KNN\\ Number of neighbors = 10\\ Distance metric = Euclidean\\ Distance weight = Squared Inverse\end{tabular} & $60.4\%$ & $94.6\%$ \\ \hline
Boosted Trees & \begin{tabular}[c]{@{}l@{}}Method = AdaBoost\\ Maximum number of splits = 20\\ Number of learners = 30\\ Learning rate = 0.1\end{tabular} & $62.1\%$ & $95.5\%$ \\ \hline
Bagged Trees & \begin{tabular}[c]{@{}l@{}}Maximum number of splits = $68\,880$\\ Number of learners = 30\end{tabular} & \boldmath{$80.1\%$} & \boldmath{$97.1\%$} \\ \hline
\end{tabular}
\end{table*}

\section{Results and Discussion}

For training and validation, the balanced dataset was randomly split using 10-fold cross-validation \cite{refaeilzadeh2009cross}, while the imbalanced dataset was randomly split using the holdout technique (training: $90\%$, validation: $10\%$).\par
In this work, the performance of commonly used ML methods was tested in terms of their accuracy, precision, and recall \cite{davis2006relationship} (see Table \ref{tab1}). Accuracy in this case is defined as the number of correct predictions divided by the total number of predictions, and can be represented by the following:
\begin{equation}
\label{eq1}
\text{Accuracy}=\frac{\text{TP}+\text{TN}}{\text{TP}+\text{TN}+\text{FP}+\text{FN}}
\end{equation}
where TP, TN, FP, and FN are the number of true positives, true negatives, false positives, and false negatives, respectively. Furthermore, precision and recall can be defined as:
\begin{equation}
\begin{split}
\label{eq2}
\text{Precision}&=\frac{\text{TP}}{\text{TP}+\text{FP}}\\
\text{Recall}&=\frac{\text{TP}}{\text{TP}+\text{FN}}.
\end{split}
\end{equation}

As shown in Table \ref{tab1}, the best performing model was the bagged trees for both balanced and imbalanced datasets. It achieved $80.1\%$ accuracy using a balanced dataset and $97.1\%$ accuracy using the imbalanced dataset. We note that the baseline accuracy of the balanced dataset is $33.3\%$ since all categories of the balanced dataset have an equal number of observations. The baseline accuracy can be achieved by a simple model that randomly allocates S4 values to either one of the three categories. Furthermore, the baseline accuracy of the imbalanced dataset is $95.46\%$, corresponding to the percentage of observations of class 1 in the imbalanced dataset, and is achieved by assuming that all predictions belong to class 1. Therefore, it is clear that the accuracy improvement gained by the model utilizing the balanced dataset is considerably higher than the baseline, by around $46.8\%$.\par

To improve the performance of the bagged trees model, Bayesian optimization was used to tune its hyperparameters, namely, the maximum number of splits and the number of learners. The optimal hyperparameters for each dataset are presented in Table \ref{tab2}.\par

\begin{table}
\centering
\caption{Optimal hyperparameters after 50 iterations of Bayesian optimization.}
\label{tab2}
\begin{tabular}{|l|c|c|}
\hline
Optimized Variables & \multicolumn{1}{l|}{\begin{tabular}[c]{@{}l@{}}Balanced\\ Dataset\end{tabular}} & \multicolumn{1}{l|}{\begin{tabular}[c]{@{}l@{}}Imbalanced\\ Dataset\end{tabular}} \\ \hline
\textbf{Maximum Number of Splits} & $378$ & $485$ \\ \hline
\textbf{Number of Learners} & $68\,828$ & $537\,360$ \\ \hline
\end{tabular}
\end{table}

\begin{table}
\centering
\caption{Confusion matrix for balanced dataset, using optimized Bagged Trees.}
\label{tab3}
\begin{tabular}{|cc|ccc|c|}
\hline
\multicolumn{2}{|c|}{\multirow{2}{*}{}} & \multicolumn{3}{c|}{\textbf{Ground Truth}} &  \\ \cline{3-6} 
\multicolumn{2}{|c|}{} & \multicolumn{1}{c|}{Class 1} & \multicolumn{1}{c|}{Class 2} & Class 3 & \emph{Precision} \\ \hline
\multicolumn{1}{|c|}{\multirow{3}{*}{\rotatebox[origin=c]{90}{\textbf{Predict.}}}} & Class 1 & \multicolumn{1}{c|}{$19\,225$} & \multicolumn{1}{c|}{$2\,076$} & $276$ & $89.11\%$ \\ \cline{2-6} 
\multicolumn{1}{|c|}{} & Class 2 & \multicolumn{1}{c|}{$3\,336$} & \multicolumn{1}{c|}{$17\,063$} & $2\,780$ & $73.61\%$ \\ \cline{2-6} 
\multicolumn{1}{|c|}{} & Class 3 & \multicolumn{1}{c|}{$398$} & \multicolumn{1}{c|}{$3\,820$} & $19\,903$ & $82.51\%$ \\ \hline
\multicolumn{1}{|c|}{} & \emph{Recall} & \multicolumn{1}{c|}{$83.74\%$} & \multicolumn{1}{c|}{$74.32\%$} & $86.69\%$ & \begin{tabular}[c]{@{}c@{}}\emph{Accuracy}\\ $81.58\%$\end{tabular} \\ \hline
\end{tabular}
\end{table}

\begin{table}
\centering
\caption{Confusion matrix for the imbalanced dataset, using optimized Bagged Trees.}
\label{tab4}
\begin{tabular}{|cc|ccc|c|}
\hline
\multicolumn{2}{|c|}{\multirow{2}{*}{}} & \multicolumn{3}{c|}{\textbf{Ground Truth}} &  \\ \cline{3-6} 
\multicolumn{2}{|c|}{} & \multicolumn{1}{c|}{Class 1} & \multicolumn{1}{c|}{Class 2} & Class 3 & \emph{Precision} \\ \hline
\multicolumn{1}{|c|}{\multirow{3}{*}{\rotatebox[origin=c]{90}{\textbf{Predict.}}}} & Class 1 & \multicolumn{1}{c|}{$377\,197$} & \multicolumn{1}{c|}{$7\,969$} & $400$ & $97.83\%$ \\ \cline{2-6} 
\multicolumn{1}{|c|}{} & Class 2 & \multicolumn{1}{c|}{$1\,627$} & \multicolumn{1}{c|}{$7\,460$} & $715$ & $76.11\%$ \\ \cline{2-6} 
\multicolumn{1}{|c|}{} & Class 3 & \multicolumn{1}{c|}{$64$} & \multicolumn{1}{c|}{$288$} & $1\,180$ & $77.02\%$ \\ \hline
\multicolumn{1}{|c|}{} & \emph{Recall} & \multicolumn{1}{c|}{$99.56\%$} & \multicolumn{1}{c|}{$47.47\%$} & $51.42\%$ & \begin{tabular}[c]{@{}c@{}}\emph{Accuracy}\\ $97.21\%$\end{tabular} \\ \hline
\end{tabular}
\end{table}

In Tables \ref{tab3} and \ref{tab4} we present the confusion matrices of the optimized bagged tree model using the balanced and imbalanced datasets, hereinafter referred to as the balanced and imbalanced models, respectively. The accuracy improvement of the balanced model after optimizing its hyperparameters was higher, by around $1.4\%$, while the accuracy of the imbalanced model improved by about $0.1\%$.\par

The overall accuracy of the imbalanced model was higher by $15.63\%$, compared to the balanced model. However, this is because the imbalanced model is biased towards the class with the higher number of observations \cite{brodersen2010balanced}. This is evident because the class 1 recall and precision of the imbalanced model were higher by $15.82\%$ and $8.72\%$, respectively, compared to the balanced model. The downside of this is the comparatively poor class 3 classification accuracy of the imbalanced model, as the class 3 recall and precision are $35.27\%$ and $5.49\%$ lower, respectively, compared to the balanced model. For class 2, the precision was $2.5\%$ higher for the imbalanced dataset, while the recall was $26.85\%$ higher for the balanced dataset. Therefore, the two models present the user with two different choices. If the user prioritizes the correct classification accuracy of weak scintillation (class 1) then the imbalanced model would be optimal. Otherwise, the balanced model would be preferable.\par

\section{Conclusion}
In this work, we attempted to solve the problem of forecasting the severity of spatially and temporally allocated amplitude scintillation values using machine learning. Six different models were tested, with the best performing of them being the bagged trees model. Next, Bayesian optimization was used to tune the hyperparameters and improve the classification accuracy. The bagged tree model trained using a balanced dataset achieved good accuracy for all three classes. On the other hand, the bagged tree model trained using an imbalanced dataset achieved excellent accuracy that was biased towards the classification of weak scintillation. We aim to expand the study by testing the models with data from different stations in different regions. Furthermore, the usage of deep learning models will be explored in future work as they may benefit from the large quantity of training data.\par

\bibliographystyle{IEEEbib}
\bibliography{Template}

\end{document}